%% file: main.tex
\documentclass[9pt,arxiv,red,secnum]{lapreprint}

\usepackage[version=4]{mhchem} 
\usepackage{siunitx}    
\usepackage{pdflscape}  
\usepackage{rotating}   
\usepackage{textgreek}  
\usepackage{gensymb}    
\usepackage[misc]{ifsym} 
\usepackage{orcidlink}  
\usepackage{listings}   
\usepackage{colortbl}   
\usepackage{tabularx}   
\usepackage{longtable}  
\usepackage{subcaption}
\usepackage{multirow}
\usepackage{snotez}     
\usepackage{csquotes}   
\usepackage{annotate-equations}
\DeclareSIUnit\Molar{M}

\usepackage[			
	backend=biber,      
    style=apa,   
	natbib=true,		
	hyperref=true,	    
	alldates=year,      
    uniquename=false,   
]{biblatex}

\AtEveryBibitem{
    \clearfield{urlyear}
    \clearfield{urlmonth}
    \clearlist{language}
}

\title{Using eye tracking to investigate what native Chinese speakers notice about linguistic landscape images}

\author[ \orcidlink{0009-0008-7739-0648} 1,2 \Letter]{Zichao Wei}
\author[1,2\Letter]{Yewei Qin}

\affil[1]{Wuhan University}
\affil[2]{School of Chinese Language and Literature}

\metadata
  []{\Letter\hspace{.5ex} For correspondence}
  {\href{mailto:exusiai.wei@outlook.com}{Zichao Wei}, \href{mailto:qyewei@gmail.com}{Yewei Qin}}
\metadata[]{Data availability}{Data availability statement. Preprocessed data could be available on \href{https://github.com/exusiaiwei/project-eyetracking-linguistic-landscape/}{Github}.}
\metadata[]{Funding}{This research received no specific funding.}
\metadata[]{Competing interests}{The author declare no competing interests.}

\leadauthor{Zichao Wei}
\shorttitle{Using eye tracking to investigate what native Chinese speakers notice about linguistic landscape images}

\begin{document}
\maketitle
\input{src/chapters/abstract}
\input{src/chapters/introduction}
\input{src/chapters/ExperimentalDesign}
\input{src/chapters/DataAnalysis}
\input{src/chapters/discussion}
\input{src/chapters/Conclusion}
\nocite{*}
\printbibliography

\if@endfloat\clearpage\processdelayedfloats\clearpage\fi



\end{document}

%% file: src/chapters/abstract.tex
\begin{abstract}

Linguistic landscape is an important field in sociolinguistic research. Eye tracking technology is a common technology in psychological research. There are few cases of using eye movement to study linguistic landscape. This paper uses eye tracking technology to study the actual fixation of the linguistic landscape and finds that in the two dimensions of fixation time and fixation times, the fixation of native Chinese speakers to the linguistic landscape is higher than that of the general landscape. This paper argues that this phenomenon is due to the higher information density of linguistic landscapes. At the same time, the article also discusses other possible reasons for this phenomenon.

\section*{Keywords}
linguistic landscape, eye tracking, fixation time, fixation times

\end{abstract}

%% file: src/chapters/introduction.tex
\section{Introduction}
The linguistic landscape is an important area of study in sociolinguistics, focusing on the language signs and language phenomena presented in public spaces. The concept of linguistic landscape, as defined by \textcite{1997-Landry-Linguistic-JoLaSP}, refers to "the visibility and salience of languages on public and commercial signs in a given territory or region. It is proposed that the linguistic landscape may serve important informational and symbolic functions as a marker of the relative power and status of the linguistic communities inhabiting the territory."

From a research perspective, current studies mainly focus on the linguistic landscape as a social symbol, including power relations, social identities, and language policies. There are relatively few studies that explore the process of linguistic landscape perception and understanding by language users.

In terms of research methods, traditional linguistic landscape studies adopt a combination of qualitative and quantitative approaches in applied linguistics research. Influenced by the research perspective, qualitative analysis is primarily used for research purposes, with a small portion of quantitative methods involved in data collection and organization, although these tend to be relatively simple. The limitations of qualitative analysis and simple data collection techniques make it difficult for researchers to focus on the actual attention process of the public towards the linguistic landscape. 

However, eye tracking technology, commonly used in psycholinguistic research, allows researchers to capture real-time information on how individuals view stimuli. Eye tracking technology has been widely applied in areas such as text reading, advertising psychology, and urban landscapes, leading to valuable insights. However, in linguistic landscape research, only a few studies have employed this innovative technique.

Currently, there are few studies on linguistic landscape that utilize eye-tracking methods. To our knowledge, there are only two known examples. \textcite{2015-Seifi-Eye-} conducted an eye-tracking study on the multilingual linguistic landscape in Leuven, Netherlands. The study found that participants spent the most time and attention on posters, inscriptions, awnings, and store names displayed in shop windows. Among monolingual and multilingual signs, the order of decreasing attention and total fixation time was as follows: Dutch, Dutch-English, English, and English-other languages. The study suggested that the location and language of signs influence people's attention towards them.

\textcite{2017-Vingron-Using-LL} collected bilingual English-French linguistic landscape data in Montreal, Canada. They investigated differences in gaze patterns of bilingual individuals when observing real linguistic landscape images containing texts in only their first or second language, or a mix of both. The study revealed that, in monolingual signs, participants consistently paid more attention to the textual part, particularly in the early stages of the experiment, with a higher fixation count on the text compared to other parts. In multilingual signs, bilingual individuals tend to initially focus on the most prominent text (in French).

Some monographs also discuss the use of eye-tracking in linguistic landscape research, such as \textcite{2020-Malinowski-Reterritorializing-}and \textcite{2021-Blackwood-Multilingualism-}. However, these monographs also discuss the case study of \textcite{2017-Vingron-Using-LL}, which focuses on the use of eye-tracking in the analysis of linguistic landscapes.

In the field of adjacent research areas such as print advertisements, newspaper layouts, and urban landscapes, there are quite a few studies that have utilized eye-tracking.

Regarding the issue of eye fixation on text and images, research in the domain of print advertisements has shown that readers tend to focus more on text than on images. \textcite{2001-Rayner-Integrating-JEPA} conducted an experiment where participants read full-page color print advertisements and found that participants spent significantly more time on the textual portions than on the visual elements. \textcite{2002-Pieters-Breaking-MS} discovered in their study that copy elements in advertisements in commercial magazines received the most attention, while brand and visual elements in retail magazines attracted more fixations. \textcite{2009-Judd-Learning-2I1ICCV} built an eye-tracking database using 1,003 images from 15 audiences and trained a saliency model with machine learning to predict fixation locations. 

The results showed that during free viewing, participants consistently paid more attention to text, humans, and faces. \textcite{2016-QianLi-Visual-TM} investigated gaze patterns of consumers on travel photos that contained text naturally embedded in the scenery. They found that regardless of whether participants understood the language of the text, textual information in the travel images attracted more visual attention. \textcite{2017-Vingron-Using-LL}, particularly focusing on linguistic landscapes, presented data showing that fixation on text was consistently higher than fixation on images, regardless of the language presented in the linguistic landscape. Furthermore, the study provided line graphs illustrating changes in fixation patterns over time.

Based on the aforementioned studies, the current research focuses on the eye fixation patterns of Mandarin native speakers when observing linguistic landscapes, specifically examining their attention to the textual and visual components. The objective is to investigate whether the observation patterns of Mandarin native speakers regarding linguistic landscapes are similar to those found by \textcite{2017-Vingron-Using-LL} with English or French native speakers. Through this cross-cultural comparison, the aim is to explore universal underlying patterns in human perception of linguistic landscapes.

%% file: src/chapters/ExperimentalDesign.tex
\section{Experimental Design}
\subsection{Participants}

Nine participants were selected for this experiment. They were all college students aged between 18 and 22 years old. They were native Chinese speakers with uncorrected or corrected visual acuity of 1.0 or above and normal color vision. They were familiar with the landscape environment of the photo shooting locations and had the ability to read the textual information in the linguistic landscape.

\subsection{Eye-Tracking Equipment}

Early eye-tracking techniques used direct observation, after-image, mechanical recording, and other methods, which had various limitations and have been eliminated in the research field. Contemporary eye-tracking methods can be divided into electro-oculography, electromagnetic induction, image or video-based eye tracking, and video recording based on pupil and corneal reflection. Among them, video recording based on pupil and corneal reflection is more prevalent, and many studies have used eye-tracking devices based on this method.

In this study, the experimental equipment chosen was the SR Research Eyelink 1000+ eyes tracker. The sampling rate was set at 1000 Hz.

The working principle of the Eyelink 1000+ eyes tracker involves using a camera connected to a computer to emit infrared light towards the eyes. The infrared light produces measurable reflections on the front and back of the cornea, while the pupil does not reflect light. By measuring the reflections on the pupil and cornea, researchers can accurately estimate eye movements when observing objects.
\subsubsection{Experimental Stimuli}
The experimental stimuli consist of 39 images, including 41 real-life storefront photos taken from shopping centers located within Wuhan city. Invalid images caused by exposure and improper focus were eliminated, resulting in a total of 39 valid materials. Each photo contains text content, mainly consisting of storefront signs, as well as a few advertisements and banners with text content within the scene layout. Among them, the proportion of different language texts in the visual landscape is shown in Table \ref{tab:language_proportion} below:
\begin{table}[ht]
    \centering
    \caption{Proportion of Different Language Texts in Visual Landscape}
    \label{tab:language_proportion}
    \begin{tabular}{cccccc}
        \toprule
        & Pure Chinese & Pure English & Chinese-English Mix & Chinese and Other Languages \\
        \midrule
        Count & 21 & 2 & 14 & 2 \\
        Proportion & 53.85\% & 5.13\% & 35.90\% & 5.13\% \\
        \bottomrule
    \end{tabular}
\end{table}

It can be seen that the text on the signs is mainly in Chinese, with some signs being a mix of Chinese and English. Pure English signs and signs with other languages have a relatively low proportion. 

This is in line with the regulations of the "Law of the People's Republic of China on the Standard Spoken and Written Chinese Language" which states that "signs and advertising words should be based on the standard spoken and written Chinese language," and it also reflects the current status of English as an internationally dominant language.
\subsubsection{Experimental Hypotheses}
Based on the research findings of \textcite{2017-Vingron-Using-LL}, we hypothesize the following in the context of the Chinese linguistic landscape:

In a Chinese language environment, native Chinese speakers will have longer gaze durations on linguistic landscapes compared to general landscapes.
In a Chinese language environment, the trend of gaze frequency on linguistic landscapes among native Chinese speakers will follow the same pattern as previous studies, with a greater focus on linguistic landscape components initially.
\subsubsection{Experimental Procedure}
The entire experiment was conducted using the SR-Research Eyelink+ eye tracker, recording eye movements at a rate of 1000Hz. Prior to the start of the experiment for each participant, the eye tracker was calibrated and validated to ensure an average calibration error of less than 0.5 degrees and a maximum calibration error of less than 1 degree. \footnote{This is the criteria for the experimental software to be considered good.}

All experimental materials were imported into the experimental program in a random order. At the beginning of the experiment, appropriate instructions were provided. During the experiment, participants viewed the images sequentially, with each image presented for 8 seconds.

%% file: src/chapters/DataAnalysis.tex
\section{Data Analysis}
\subsection{Interest area}
We use the DataViewer software, developed by SR-Research, to analyze and process the data. First, interest areas (IA) are created within each experimental image to calculate relevant metrics within these areas. 

An interest area (IA) refers to a specific region selected by the experimenters, with its position and range pre-defined. In the experiment, there are two types of IAs for each image. 

The IA surrounding the text portion is labeled as "T" (text), while the IA surrounding the image portion is labeled as "I" (image). Once the IA labeling is completed, we select the start and end times of each fixation point, along with the IA type of the fixation point, export the relevant data, and proceed with subsequent analysis.
\subsection{Fixation Time}
\subsubsection{Average Duration of Single Fixation}
In all experiments, a total of 9338 fixation data were collected, with 4885 fixations on the text portion and 3200 fixations on the image portion. There were 1253 invalid data points located outside the interest areas. The average duration of a single fixation on the text portion was approximately 261.13ms, while for the image portion, it was approximately 244.99ms. 
\subsubsection{Average Fixation Time per Participant per Image}
In the experiment, on average, each participant spent 3634.19ms fixating on the text portion of each image and 2233.50ms fixating on the image portion. A univariate analysis of variance showed that the fixation duration on the text portion was significantly greater than that on the image portion (p<0.01). The Levene's test for equality of error variances yielded a significance value > 0.05, indicating that the assumption of equal variances for the T and I groups is not rejected and the significance data is valid.
\subsubsection{Hypothesis Conclusion }
Based on the aforementioned analysis, we conclude that hypothesis 1 is supported.
\subsection{Fixation Count}
\subsubsection{The Variation of Fixation Counts and Proportion Over Time}
To determine whether the participants were fixating on text or non-text objects at specific time points, we sampled data every 10ms from 10ms to the end of data recording at 8000ms, resulting in a total of 800 coordinate points. If a coordinate point fell within the duration of a fixation, it was counted. The analysis results were visualized as a line graph to observe the trend.

\begin{figure}[ht]
    \centering
    \includegraphics[width=0.8\textwidth]{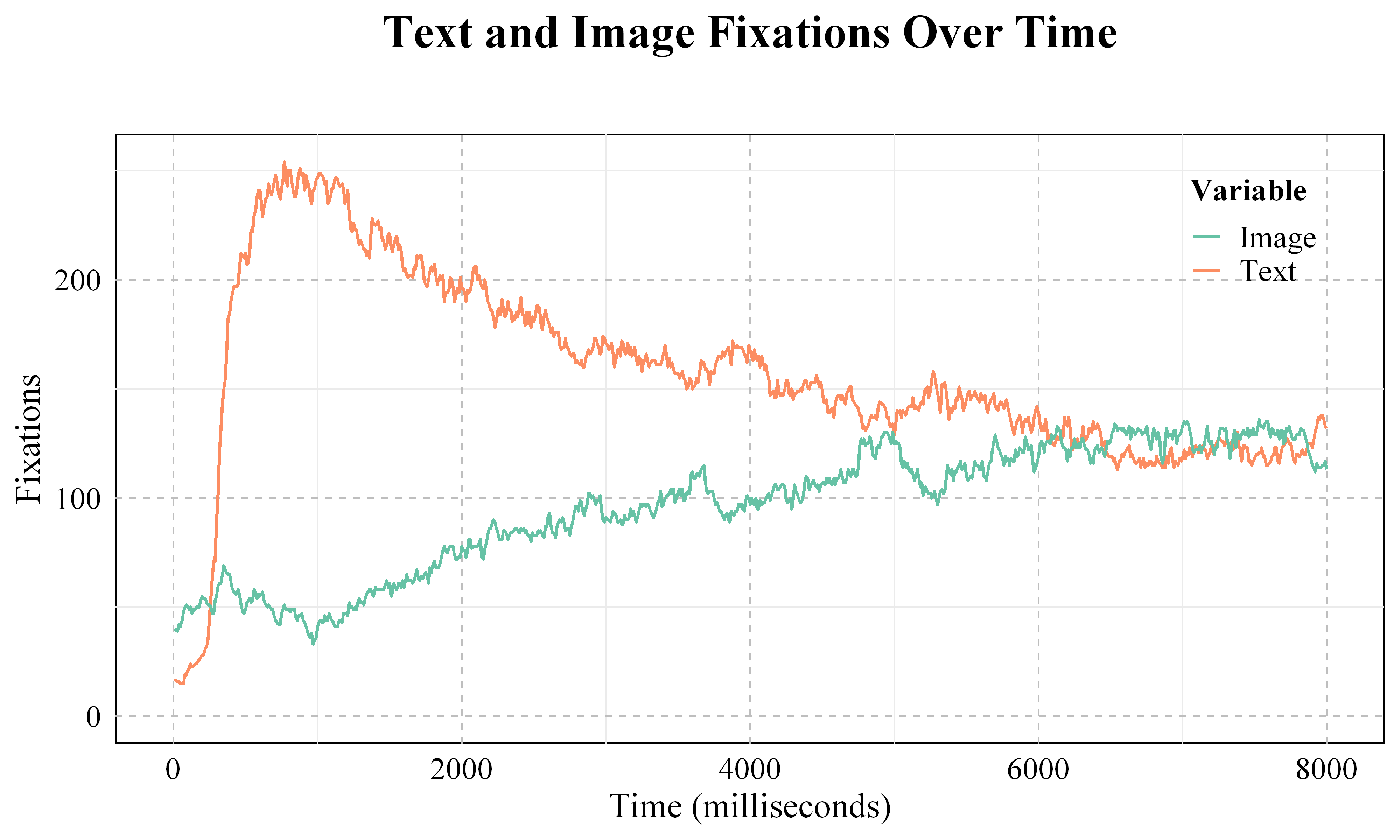}
    \caption{Changes in Fixations on Text and Images Over Time}
    \label{fig:fixations}
\end{figure}

\begin{figure}[ht]
    \centering
    \includegraphics[width=0.8\textwidth]{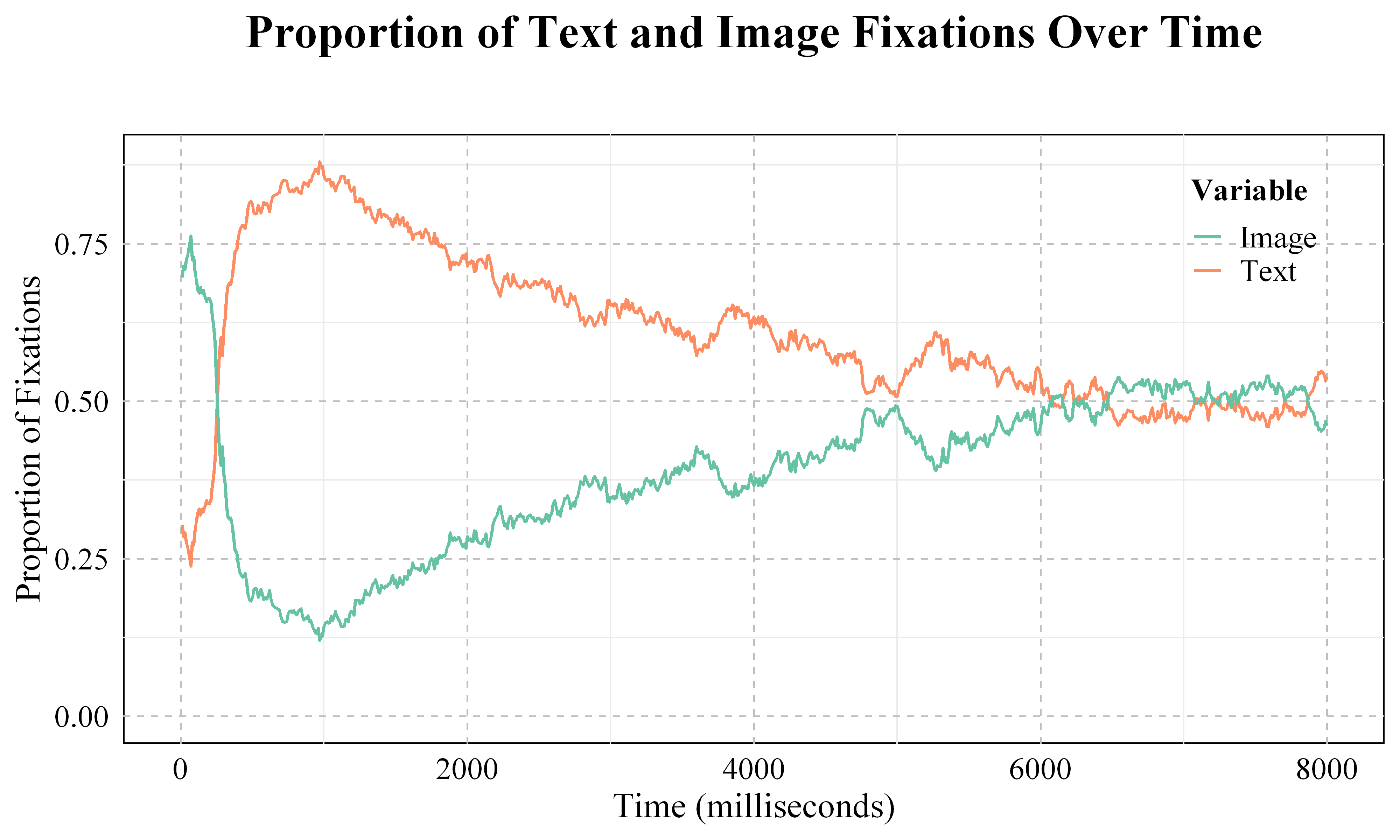}
    \caption{Proportion of Fixations on Text and Images Over Time}
    \label{fig:proportion}
\end{figure}

We created two line graphs depicting the changes in the fixations on text and images over time (Figure \ref{fig:fixations}) and the proportion of fixations on text and images over time (Figure \ref{fig:proportion}).

From the line graphs, it can be observed that overall, there were more fixations on the text compared to the images. At around 250ms, there were relatively fewer fixations on both text and images, with slightly more fixations on the images at this point. Subsequently, within the interval of approximately 250ms to 1000ms, the number of fixations on the text rapidly increased and reached its peak at around 1000ms. Afterward, the number of fixations on the text started to decrease, with some slight rebounds around 3500ms-4200ms and 5000ms-6000ms, forming two new peaks. Towards the end of the experiment, the fixations on the text continued to decrease while the fixations on the images continued to increase, with the phenomenon of slightly more fixations on images than on text appearing for the first time in this period.

The overall trend of fixation patterns reflected in Figure \ref{fig:proportion} is similar to Figure \ref{fig:fixations}. In terms of proportions, the number of fixations on text is consistently higher than on images. 

However, there is a difference in that at around 250ms, when fixations on both text and images are relatively low, the proportion of fixations on images is significantly higher than on text.

Overall, the changes in fixation counts over time exhibit three characteristics: 1) more fixations on text than on images, 2) a gradual decrease in the proportion of fixations on text and an increase in the proportion of fixations on images, approaching consistency in the later stage, and 3) in the early stage of the experiment (before approximately 250ms), slightly more fixations on images than on text, and a much higher proportion of fixations on images compared to text.

\subsubsection{The Number of Fixations in Different Time Periods}
In order to analyze the fixation patterns in different time periods, we adopted the time interval division method proposed by \textcite{2017-Vingron-Using-LL}. We divided the total experimental duration of 8000ms into three time periods: the first time period was from 0-2666ms, the second time period was from 2667-5333ms, and the third time period was from 5334-8000ms. Using paired samples t-tests, we compared whether there were significant differences in the number of fixations allocated to each time period.

The statistics of the number of fixations in each period are shown in the following table \ref{tab:fixations}:
\begin{table}[ht]
    \centering
    \caption{Fixations on Text and Images Over Different Periods}
    \label{tab:fixations}
    \begin{tabular}{cccc}
        \toprule
        & Period 1 & Period 2 & Period 3 \\
        \midrule
        Text & 50469 & 41420 & 34060 \\
        Image & 16346 & 27526 & 33102 \\
        \bottomrule
    \end{tabular}
\end{table}
Paired samples t-tests revealed that there were significant differences in the number of fixations on text between the first and second time periods (t=8.646, p<0.01), as well as between the second and third time periods (t=43.748, p<0.01).
\subsubsection{Hypothesis Conclusion}
Based on the above analysis, we conclude that hypothesis 2 is supported. However, further discussion is needed to explore the differences between the current data and previous studies.

%% file: src/chapters/discussion.tex
\section{Discussion}
\subsection{Comparison}
\subsubsection{Comparison with \textcite{2017-Vingron-Using-LL}}
The changes in the number of annotations over time in this study are generally consistent with the gaze patterns of native linguistic landscapes described by \textcite{2017-Vingron-Using-LL}. In terms of overall trends, the number of gazes on text is greater than that on objects, and the number of gazes on text is higher in the first period than in the second and third periods. From the maximum value of gaze ratio, the gaze ratio on text reaches its peak at around 1000ms in the first period. Subsequently, there are two to three small rebounds in the gazes on text, represented by a large peak followed by two to three smaller waves on the image.

During the experiment, as time goes on, the number of gazes tends to regress towards the image. However, there are still some deviations between the two experiments in terms of specific details. In \textcite{2017-Vingron-Using-LL}, regardless of the type of image and the language background of the subjects, the number of gazes on text throughout the experiment is higher than that on objects. 

However, in this study, in the early stage of the first period at around 250ms, the number of gazes on the image overwhelms that on the text, and in the regression of the third period, there is a brief situation where the gazes on the image are slightly higher than those on the text. In \textcite{2017-Vingron-Using-LL}, although the gaze ratio on the text in the first period is higher than that in the third period, the significance of this difference is small (p = 0.065>0.05), while in this study, there is a highly significant difference in this difference (p <0.01).
\subsubsection{Comparison with Studies on Print Advertisements }
The data from this study indicate that in complex real environments, the total gaze time and single gaze time on the text part of the linguistic landscape are significantly longer than the gaze time on the image part. This conclusion is similar to the findings of \textcite{2001-Rayner-Integrating-JEPA}, but there are some differences compared to studies on print advertisements, such as the conclusions of \textcite{2002-Pieters-Breaking-MS}

We believe that although both print advertisements and linguistic landscapes in real scenes contain text and images as the research objects, they differ significantly in terms of the organization of content, the amount of information contained, and the environment they are in. For example, print advertisements have a smaller area and are read at a closer distance, while linguistic landscapes have a larger area and people observe them from a greater distance. For print advertisements, researchers can adjust the arrangement of their constitutive factors based on experimental designs to conduct controlled variable studies. However, linguistic landscapes come from real language life scenes, and researchers cannot and should not modify their content. Linguistic landscapes often exist in more complex real environments, influenced by various external factors, while print advertisements are relatively independent, further reducing potential interference. 

Therefore, we believe that there may be similar or identical cognitive mechanisms involved in the gaze on linguistic landscapes and print advertisements. However, considering the differences in research objects, the data contradicting the findings of this study in previous research are difficult to negate the conclusions of this study.
\subsection{Analysis}
\subsubsection{Research Material: Complex Real Scenes}
We found that the images used in this study have higher complexity compared to the experimental materials in previous related research, mainly due to the complexity of real scenes. 

Compared to print advertisements, urban landscapes occupy a larger space and contain broader information. Advertisements may only be a part of store design. At the same time, print advertisements are static images, while the constantly changing urban landscape is more similar to a complex system. For example, print advertisements remain unchanged after being created, while storefront landscapes may change due to marketing activities, the wishes of store owners or customers, and requirements from authorities. Therefore, the evolving urban landscape has greater complexity and possibilities. 

Compared to \textcite{2017-Vingron-Using-LL}, the research material in this study is also more complex. According to the experiment materials disclosed in their paper, we can see that the study focuses on linguistic landscapes such as road signs and billboards, where both textual and image information are relatively limited, and the overall amount of information in the picture is also less compared to the images selected from shopping centers in major cities in this study. 

We speculate that the phenomenon of higher gazes on the image part than the text part in the first 250ms, as observed in this study, may be due to differences in research materials. Faced with real scene images, participants may tend to gather more information about the overall scene through observation at the beginning, in order to confirm their location and facilitate subsequent actions. 

However, studies on print advertisements and \textcite{2017-Vingron-Using-LL} may not show this phenomenon due to the nature of their research materials. \textcite{2016-QianLi-Visual-TM} studied images of real tourist scenes but did not provide information on gaze changes over time. This hypothesis still requires further validation in future research.
\subsubsection{Information Density}
Research on scene perception has found that participants tend to fixate on areas that contain more "information" or areas that they consider important \parencite{1935-Buswell-How-,1967-Yarbus-Eye-EMV}. This conclusion is also reflected in studies on print advertising, where consumers pay more attention to the areas they are prompted to focus on \parencite{2001-Rayner-Integrating-JEPA}. Scene perception research has proposed two hypotheses to explain this phenomenon: the semantic information hypothesis and the perceptual factors hypothesis. 

The study by \textcite{1967-Mackworth-Gaze-P&P} revealed that participants fixate on areas with higher information density during the early stages of scene viewing, and as time progresses, they maintain a high proportion of fixations on high information density regions. \textcite{1974-Antes-Time-JEP} conducted a similar study using complex images and found that participants also fixate on areas with higher information density in the early stages, but what differs from previous research is that participants tend to look more at the parts of the picture with lower information density in the later stages. The conclusions of these studies are consistent with the findings of this article, which suggest that in urban landscapes, textual components have higher information density compared to pictorial components, which is consistent with the communicative function of language. 

However, according to this theory, how can we explain the findings of studies such as \textcite{2002-Pieters-Breaking-MS}, which indicate that participants fixate more on pictures than on text? We believe that this discrepancy is related to the differences between real-world complex scenes and advertising materials. Advertising text is designed to attract consumers, and due to the limited space in print ads, designers must make use of every inch to convey information. 

Carefully designed images can convey more information than text, such as the presentation of product appearance. In real-world scenes, both the top-down and bottom-up linguistic landscapes have been encoded by designers, and their textual content is as concise as the components in print ads. 

However, the image components, such as store decorations, arrangements, and layouts, are limited by the physical properties of public spaces and have a smaller design space. Even though there are design factors involved, they can only indirectly convey information. At the same time, the image components in the linguistic landscape occupy a larger area, which dilutes the information contained in them, resulting in lower information density compared to the textual components. 

Therefore, the textual components of the linguistic landscape in public spaces have higher information density, leading to more fixations on the text than on the images. Images in print ads may also have higher information density, leading to the conclusion that some studies have found more fixations on pictures than on text. 

These seemingly contradictory conclusions may actually follow the same cognitive principles. The hypothesis that perceptual factors influence fixation density suggests that the fixation pattern in scenes is influenced by visual factors. Several studies have proposed visual factors that influence the initial fixation location, including maximum brightness, minimum brightness, image contrast, maximized local positive physiological comparisons, minimized local negative physiological comparisons, edge density, and high spatial frequency. 

The materials used in this study were also influenced by these factors, for example, many parts of the linguistic landscape were brightly lit. However, due to the limitations of the experiment scale, it was difficult to find enough materials to control these variables for a more detailed analysis. This perspective can be used as a reference for future researchers.
\subsubsection{Reading Time}
It is generally believed that longer fixation times on a certain part indicate more interest from participants in that part. In this study, both the average fixation time per fixation and the average fixation time per participant per image were significantly longer for the textual components of the linguistic landscape compared to the image components. This indirectly supports the hypothesis that the textual components of the linguistic landscape provide more information than the image components. 

Research on visual encoding time in scene perception by \textcite{2009-Rayner-Eye-PS} showed that scenes need to be presented for at least 150ms to ensure normal cognitive processing, which is longer than the processing time required for reading. In this study, the mean fixation time on the textual components (261.13ms) and the mean fixation time on the image components (244.99ms) were both longer than 150ms, satisfying the requirements for cognitive processing.
\subsubsection{Cultural Differences }
\textcite{2005-Chua-Cultural-PNAS} study indicated that different cultural groups have differences in the objects they focus on in scene perception. Their research found that Americans had more fixations and longer fixation times on objects in scenes compared to Chinese participants, while Chinese participants had more fixations on the background compared to Americans. This study interpreted these differences in cognitive processing as being caused by cultural differences, with Chinese culture emphasizing collectivism and American culture emphasizing individualism.

The participants in this study were all Chinese, while the participants in \textcite{2017-Vingron-Using-LL} were from Western cultural backgrounds. Although, in general, participants in this study still had more fixations on the textual components, it is worth noting that the proportion of fixations on the image components in this study was still much higher compared to the proportion of fixations on non-textual objects in \textcite{2017-Vingron-Using-LL}, with the former approaching 50$\%$ at some points, while the latter consistently remained significantly lower than the fixation proportion on text. The differences in the materials used in the studies can theoretically account for this difference, but it remains to be investigated whether there are cultural influences behind this difference. Further research is needed to delve into this issue.
\subsubsection{Informational Function of Linguistic Landscape}
After discussing the possible impact of information density on differences in gaze patterns between text and image, it is necessary to return to the field of sociolinguistics and review the information functions of linguistic landscapes. \textcite{1997-Landry-Linguistic-JoLaSP} proposed that the two basic functions of linguistic landscapes are informational and symbolic. The former can be broadly divided into three levels: linguistic landscapes as unique indicators of the geographical environment of linguistic communities; linguistic landscapes that reflect language characteristics and boundaries of a region; linguistic landscapes that provide information on the social-linguistic composition of the corresponding geographical area. However, it can be seen from the definition that the informational function of linguistic landscapes is purely social and does not involve the biological process of human reception of information while viewing linguistic landscapes.

Considering that the process of human language use (including the process of observing, using, and generating linguistic landscapes) is both social and biological, we believe that, despite the valuable studies produced from the sociolinguistic perspective of linguistic landscapes in the long-term research, introducing a biological perspective to understand the real process of humans observing and obtaining information from linguistic landscapes will bring broader research perspectives and produce more valuable results. For example, the symbolic function of linguistic landscapes inevitably involves the actual process of human observation of linguistic landscapes. The difference observed in \textcite{2017-Vingron-Using-LL} between English native speakers and French native speakers when observing mixed English-French linguistic landscapes is favorable evidence for research on the symbolic function of linguistic landscapes.

Therefore, we believe that it is necessary to update the definition of the informational function of linguistic landscapes and include the process of humans actually receiving information from linguistic landscapes when viewing them in the informational function of linguistic landscapes and conduct studies using psychological linguistic research methods such as eye-tracking.

%% file: src/chapters/Conclusion.tex
\section{Conclusion}
The experimental results indicate that, in the context of the Chinese language environment, native Chinese speakers have longer fixation times on the textual components of the linguistic landscape compared to the image components. They also have a higher number of fixations on the textual components compared to the image components, and this trend shows a decrease over time. 

We believe that this may be because the textual components of the linguistic landscape in real-world scenes carry more information and have a higher information density. 

Additionally, there are still many perspectives to consider regarding this issue, and we hope that further research can propose new viewpoints. Furthermore, we believe that the linguistic landscape is a complex entity in the real world, and the interpretation of linguistic landscape-related issues relies on the integration of various research paradigms. 

Eye-tracking technology provides us with new perspectives for understanding linguistic landscape issues. Incorporating this method into the toolbox of sociolinguistics, integrating it with existing research methods that have made significant contributions to linguistic landscape studies, may help us further understand language life in public spaces and provide new theoretical foundations for sociolinguistic research and language policy and planning.